%% file: main.tex
\documentclass{article} 
\usepackage{MLDD_workshop_2023,times}

\input{math_commands.tex}

\usepackage[T1]{fontenc} 
\usepackage{array}
\usepackage{multirow}
\usepackage{amsmath}
\usepackage{amsfonts}       
\usepackage{graphicx}
\usepackage{hyperref}
\usepackage{booktabs}       
\usepackage{nicefrac}       
\usepackage{microtype}      
\usepackage{xcolor}         
\usepackage{subfig}
\usepackage{wasysym}

\title{Multi-scale Sinusoidal Embeddings Enable Learning on High Resolution Mass Spectrometry Data}

\author{
Gennady Voronov,  \\
Enveda Biosciences\\
Boulder, CO 80301, USA \\
\texttt{gennady.voronov@envedabio.com}
\And
Rose Lightheart\\
Enveda Biosciences\\
Boulder, CO 80301, USA \\
\texttt{rose.lightheart@envedabio.com}
\And
Joe Davison\\
Unaffiliated\thanks{This work was completed while J.D. was an employee at Enveda Biosciences.}\\
\texttt{joeddav@gmail.com}
\And
Christoph A.~Krettler\\
Enveda Biosciences\\
Boulder, CO 80301, USA \\
\texttt{christoph.krettler@envedabio.com}
\And
David Healey\\
Enveda Biosciences\\
Boulder, CO 80301, USA \\
\texttt{david.healey@envedabio.com}
\And
Thomas Butler\\
Enveda Biosciences\\
Boulder, CO 80301, USA \\
\texttt{tom.butler@envedabio.com}
}

\iclrfinalcopy 
\begin{document}

\maketitle

\begin{abstract}
Small molecules in biological samples are studied to provide information about disease states, environmental toxins, natural product drug discovery, and many other applications. The primary window into the composition of small molecule mixtures is tandem mass spectrometry (MS2), which produces high sensitivity and part per million resolution data. We adopt multi-scale sinusoidal embeddings of the mass data in MS2 designed to meet the challenge of learning from the full resolution of MS2 data. Using these embeddings, we provide a new state of the art model for spectral library search, the  standard task for initial evaluation of MS2 data. We also investigate the task of chemical property prediction from MS2 data, that has natural applications in  high-throughput MS2 experiments and show that an average $R^2$ of 80\% for novel compounds can be achieved across 10 chemical properties prioritized by medicinal chemists. We vary the resolution of the input spectra directly by using different floating point representations of the MS2 data, and show that the resulting sinusoidal embeddings are able to learn from high resolution portion of the input MS2 data. We apply dimensionality reduction to the embeddings that result from different resolution input masses to show the essential role multi-scale sinusoidal embeddings play in learning from MS2 data.

\end{abstract}

\section{Introduction\label{sec:Introduction}}

Metabolomics is the study of the small molecule ($\apprle\!1,\!000\,\mathrm{Daltons}$) contents of complex biological samples. Tandem Mass Spectrometry (MS/MS), in conjunction with chromatography, is one of the most commonly used tools in metabolomics. Tandem Mass Spectrometry works by measuring with very high resolution the masses of molecules and their constituent fragments. While MS/MS techniques are highly sensitive and precise, inferring the identity of the molecules and their properties from the resulting mass spectra is commonly regarded as one of metabolomics' primary bottlenecks \citep{dunn2013mass}. Improved tools for these tasks will impact applications across many areas of science including disease diagnostics, characterization of disease pathways, development of new agrochemicals, improved forensics analysis, and the discovery of new drugs \citep{zhang2020mass}.

Profiling unknown molecules with mass spectrometry consists of several steps. First, molecules of interest are ionized and separated by their mass to charge ratio ($m/z$), resulting in the MS1 spectrum. Then, individual ``precursor'' ions are fragmented, and the $m/z$'s of the fragments are recorded in the same manner. The resulting spectrum contains the $m/z$'s and intensities (together, the ``peaks'') of all resulting fragments, and is called the MS2 spectrum. See \citet{glish2003basics}.


In recent years, several machine learning methods have been developed to identify the structures and properties of small molecules from their mass spectra. These approaches \citep{huber2021spec2vec,huber2021ms2deepscore,kutuzova2021bi,litsa2021spec2mol,shrivastava2021massgenie,van2016topic} historically discretize $m/z$ (via tokenization or binning). 
However, the $m/z$ values obtained in modern mass spectrometry experiments are collected with parts per million levels of resolution. There are three critical reasons to expect that modeling $m/z$ values as numeric quantities, in contrast to discretized values, is the appropriate technique. First, we know that relevant chemical information is present at the millidalton level \citep{jones2004strategies,pourshahian2008application} and discretization schemes typically strip this information away. Additionally, mass differences between peaks represent fragmentation patterns, and are therefore relevant to understanding the molecule. Learning to utilize this information from tokens is exceedingly difficult. Finally, discretization is susceptible to edge effects, where slight mass difference can map to different bins if the masses are close to the edge of a bin. In light of these considerations, we model $m/z$ as a numerical value using sinusoidal embeddings, which we hypothesize will enable us to capture information across many orders of magnitude.

In this work, we apply a numerical representation of $m/z$ that uses sinusoidal embeddings across multiple scales to retain the information content of MS2 data across its entire mass resolution. To demonstrate the ability of these embeddings to enable learning at the high resolution of MS2 experimental data, we apply them to a search task and 10 regression tasks. 

Our first test task is spectral library search \citep{stein1994optimization}. In spectral library search, spectra from unknown compounds are compared to spectra in a database to find matches using a similarity function over pairs of spectra. This task is the primary method used in standard metabolomics analyses, but is challenging because spectra for a compound vary widely with experimental conditions. We find that a similarity function based on sinusoidal embeddings achieves state of the art both for finding the exactly correct compound, and also for finding close structural analogs, which is useful for compounds not contained in spectral databases. 

We further investigate predicting chemical properties relevant to drug discovery from MS2 data and apply the same modeling approach to this task. We achieve 80\% average $R^2$ for out of sample molecules across 10 properties, which is high enough to enable first-pass filtering and selection of candidate drug molecules in high-throughput experiments based solely on spectral data. 

In each task, using sinusoidal embeddings results in a new state of the art. The confirmation of across-task improvement provides evidence that the embeddings are a general improvement, rather than task specific. To determine whether these results are due to learning from the high resolution portion of the data, we experiment with inputting MS2 data cast to half precision floating point numbers, and show that the performance noticeably degrades relative to double precision. Finally, we visualize embeddings generated with varied floating point precision MS2 inputs using UMAP \citep{mcinnes2018umap} projections and show that non-trivial high dimensional structure only emerges with sufficiently high precision input. Taken together, these results are the first clear evidence that sinusoidal embeddings enable effective learning from high mass resolution MS2 data across tasks in metabolomics.

\subsection{Related work\label{subsec:related-work}}

Modeling numerical data in terms of sinusoidal functions has a long history in many scientific fields. In machine learning, sinusoidal embeddings are most commonly used to encode the discrete positions of natural language token inputs in transformer models, which are otherwise position-agnostic \citep{vaswani2017attention}. Other work has used sinusoidal embeddings for multi-dimensional positional encoding in image recognition \citep{li2021learnable}. In mass spectrometry, sinusoidal embeddings have been used in proteomics for inferring protein sequences from mass values, either with \citep{qiao2019deepnovov2} or without \citep{yilmaz2022novo} an initial mass binning step, but without exploring their role in model performance or comparing to alternative approaches. These techniques have never been applied in the domain of metabolomics. Metabolomics differs from proteomics in that the molecules of of interest are 2 - 3 orders of magnitude lower mass and are graph structured rather than sequential as in proteins. Consequently, the tasks and challenges of modeling of small molecules are sharply divergent from those in proteomics.

For modeling $m/z$ values in the field of metabolomics, previous machine learning models have relied primarily on discretization of the continuous mass inputs. This is usually accomplished by binning $m/z$ values into fixed length vectors with peak intensity as the value for each element. Various authors have used binned representations of spectra for spectral library search \citep{huber2021ms2deepscore}, unsupervised topic modeling \citep{van2016topic}, and molecule identification \citep{kutuzova2021bi,litsa2021spec2mol}. Alternatively, masses have been tokenized via rounding for tasks such as unsupervised spectral similarity \citep{huber2021spec2vec} and molecule prediction from synthetic data \citep{shrivastava2021massgenie}. 

Other approaches rely on tokenization of $m/z$ values by assigning a molecular formula to each peak, and taking the formula as a token. Even when this method is able to uniquely identify molecular formulas for every fragment ion, it still faces a problem endogenous to tokenization: it's very difficult to reason about the relationships between discrete tokens except by pattern recognition over a very large number of training examples. \citet{bocker2008towards} addressed this difficulty by modeling the fragmentation process with a tree structure, and \citet{duhrkop2015searching} used these fragmentation trees as inputs to a molecular fingerprint prediction model that is used to assign molecules to spectra from molecular structure databases. The tree-based approaches are prohibitively computationally expensive for large volumes of data \cite{cao2021moldiscovery}.

\section{Model\label{sec:Model}}

\subsection{Base model architecture\label{subsec:Base-model-architecture}}

An MS2 spectrum, 
\begin{equation}
S=\left\{ \left(m/z,I\right)_{\mathrm{precursor}},\left(m/z,I\right)_{\mathrm{fragment}_{1}},\ldots,\left(m/z,I\right)_{\mathrm{fragment}_{N}}\right\} ,\label{eq:spectrum_def}
\end{equation}
is composed of a precursor $m/z$ and a set of $N$ fragment peaks at various $m/z$'s and intensities ($I$). Various other data are typically collected, including precursor abundance, charge, collision energy, \textit{etc}, but these are not used in this work. Transformer encoders \citep{vaswani2017attention} without the positional encoding are explicitly fully-symmetric functions and hence are ideally suited to model a set of fragmentation peaks. We therefore take our base model, $\mathrm{SpectrumEncoder}\left(S\right)$, to be a transformer encoder, whose inputs are a set of $\left(m/z,I\right)$ pairs that includes the precursor along with all of the fragment peaks. We normalize the intensities to a maximum of $1$ for fragments and assign an intensity of $2$ to the precursor. Finally, we take as output the embedding vector from the final transformer layer corresponding to the precursor input. See Figure \ref{fig:transformer}. Note, the use of transformers to study synthetic MS2 spectra has been explored in \citet{shrivastava2021massgenie}.

This flexible approach enables us to experiment with and compare various representations of MS2 data. We pursue two approaches: tokenization of $m/z$ and modeling $m/z$ as numerical values via sinusoidal embeddings. Before describing these, we need the following definition for a simple two layer feed forward MLP,
\begin{equation}
\mathrm{FF}\left(x\right)=W_{2}\mathrm{ReLu}\left(W_{1}x+b_{1}\right)+b_2.\label{eq:feed_forward}
\end{equation}
Note that each occurrence of $\mathrm{FF}$ we employ below is a separate instance with distinct weights. 

\subsubsection{Tokenized $m/z$ peak embedding}

Our first approach is to discretize $m/z$ by rounding to $0.1$ precision and treating these objects as tokens (as in an NLP context). These tokens are embedded in a dense vector space as in \citet{huber2021spec2vec,mikolov2013efficient}, via an embedding function $\mathrm{TE}$. We then construct the token peak embedding,
\begin{equation}
\mathrm{PE}_{\mathrm{token}}\left(m/z,I\right)=\mathrm{FF}\left(\mathrm{TE}\left(m/z\right)\parallel I\right),\label{eq:token-peak-embd}
\end{equation}
where $\parallel$ denotes concatenation. A diagram of $\mathrm{PE}_{\mathrm{token}}$ is shown in Figure \ref{fig:token-peak-embd}.

\subsubsection{Sinusoidal $m/z$ peak embedding}

\begin{figure}
\subfloat[\label{fig:token-peak-embd} Architecture of $\mathrm{PE}_{\mathrm{token}}$.]{\begin{centering}
\includegraphics[scale=0.4]{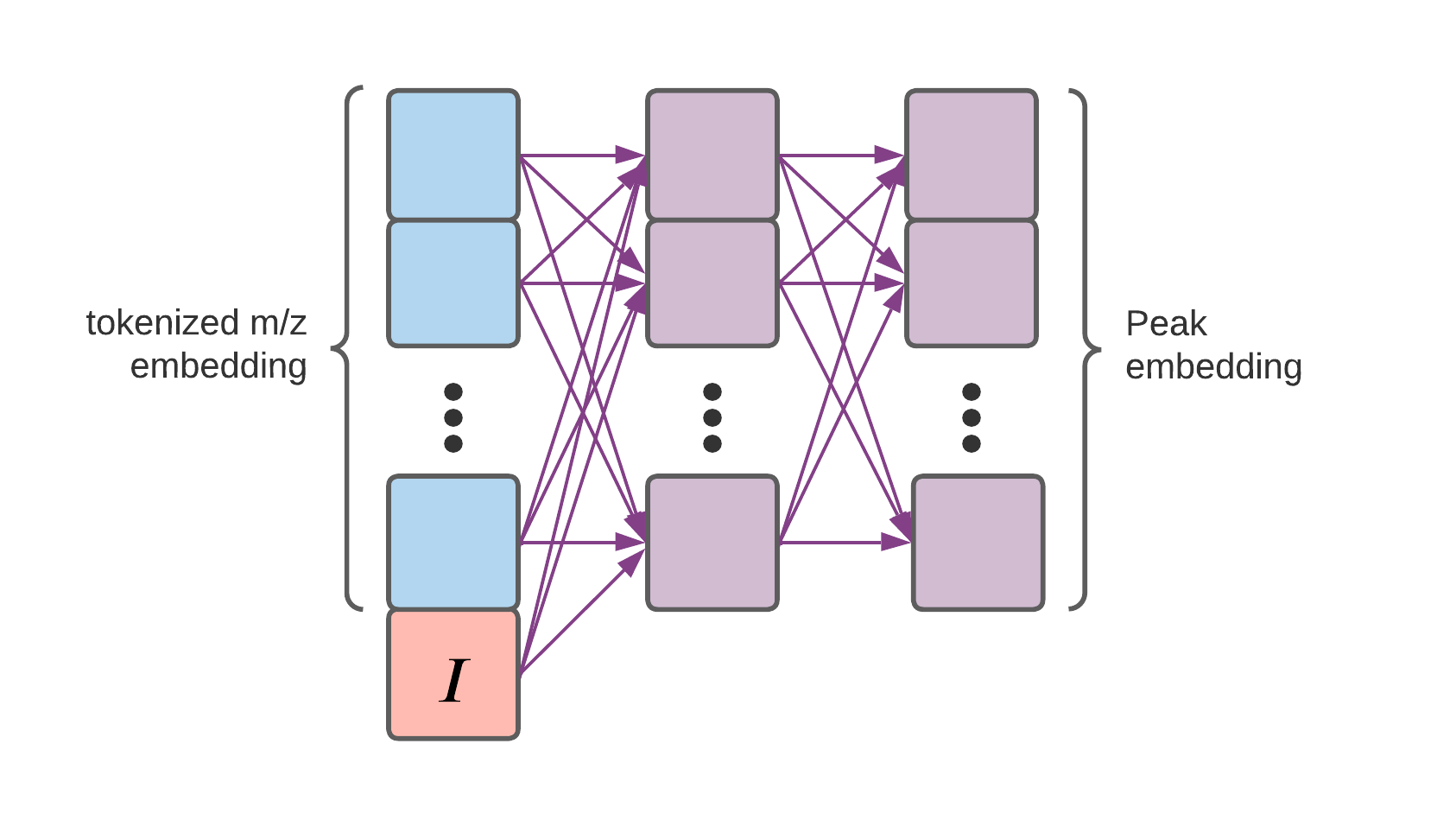} 
\par\end{centering}
}\quad{}\subfloat[\label{fig:sin-peak-embd} Architecture of $\mathrm{PE}_{\mathrm{sin}}$.]{\begin{centering}
\includegraphics[scale=0.4]{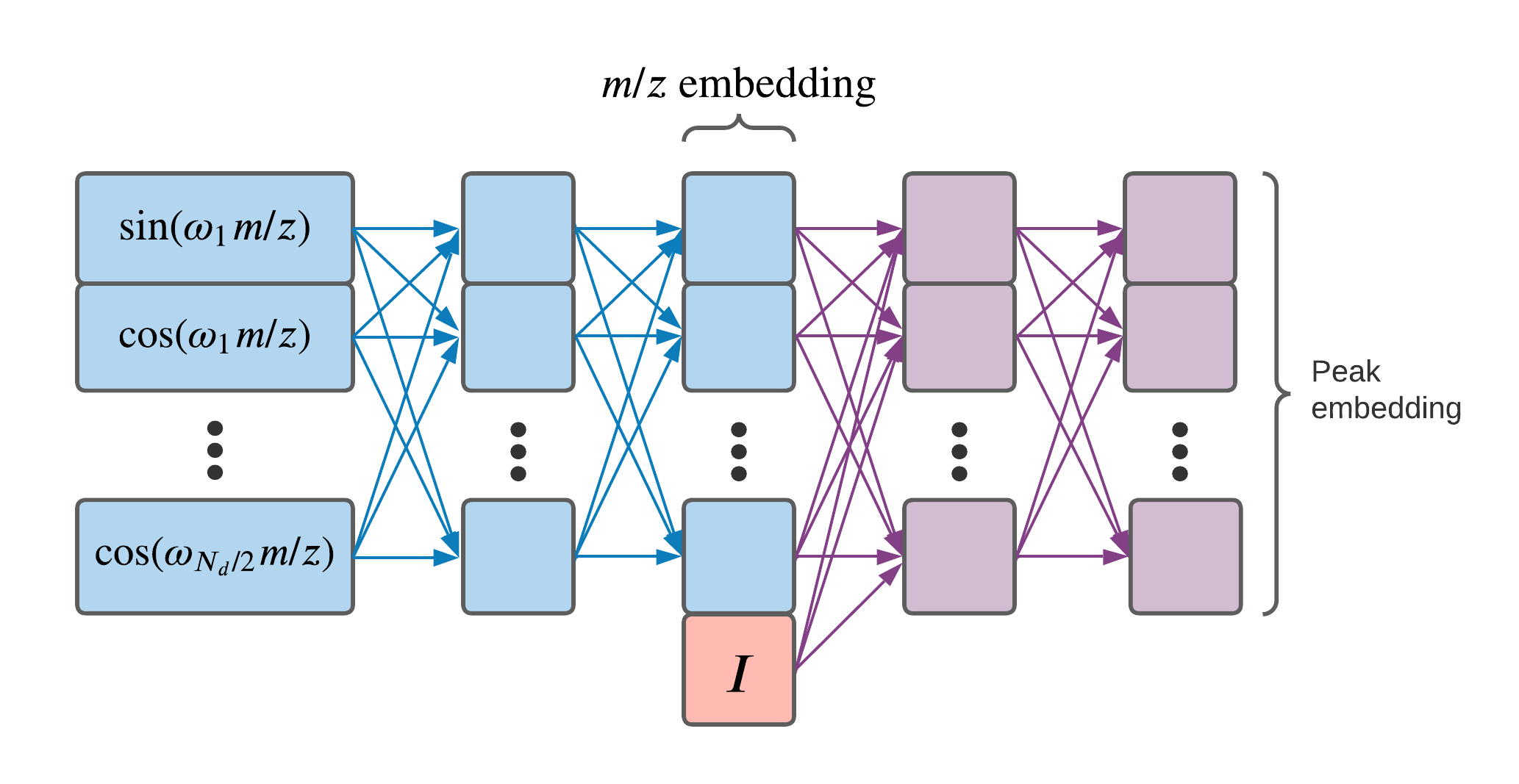} 
\par\end{centering}
}\caption{\label{fig:peak-embd} Architectures used to embed MS2 peaks (composed of a $m/z$ and an intensity $I$). Both approaches produce an embedding of $m/z$ of dimension $d$. The $m/z$ embedding in Figure \ref{fig:token-peak-embd} (left) is produced by a learned embedding layer. The $m/z$ embedding in Figure \ref{fig:sin-peak-embd} (right) is produced by a sinusoidal embedding followed by a two layer MLP. In both cases, an intensity value $I$ is appended to make a vector of dimension $d+1$. This is then fed into a simple feed-forward network to produce a peak embedding of dimension $d$. We use blue to highlight the network components involved in producing an $m/z$ embedding and purple to highlight the full peak embedding after intensity is incorporated.}
\end{figure}

We now consider a numerical peak embedding that utilizes the full $m/z$ precision. We employ a sinusoidal embedding as follows:
\begin{eqnarray}
\mathrm{SE}\left(m/z,2i;d\right) & = & \sin\left(2\pi\left[\lambda_{\mathrm{min}}\left(\frac{\lambda_{\mathrm{max}}}{\lambda_{\mathrm{min}}}\right)^{2i/\left(d-2\right)}\right]^{-1}m/z\right)\label{eq:sin_embd}\\
\mathrm{SE}\left(m/z,2i+1;d\right) & = & \cos\left(2\pi\left[\lambda_{\mathrm{min}}\left(\frac{\lambda_{\mathrm{max}}}{\lambda_{\mathrm{min}}}\right)^{2i/\left(d-2\right)}\right]^{-1}m/z\right).\label{eq:cos_embd}
\end{eqnarray}
The frequencies are chosen so that the wavelengths are log-spaced from $\lambda_{\mathrm{min}}=10^{-2.5}\,\mathrm{Daltons}$ to $\lambda_{\mathrm{max}}=10^{3.3}\,\mathrm{Daltons}$, corresponding to the mass scales we wish to resolve. This embedding is inspired by \citet{li2021learnable,vaswani2017attention}. Similarly to Equation \ref{eq:token-peak-embd}, we construct the sinusoidal peak embedding, 
\begin{equation}
\mathrm{PE}_{\sin}\left(m/z,I\right)=\mathrm{FF}\left(\mathrm{FF}\left(\mathrm{SE}\left(m/z\right)\right)\parallel I\right).\label{eq:sin-peak-embd}
\end{equation}
We also experiment with casting the $m/z$ inputs as half, single, and double precision floating point values. Equations \ref{eq:sin_embd}-\ref{eq:cos_embd} are computed at the precision specified by $m/z$ and the results are then cast to the precision of the model weights used for training. A diagram of $\mathrm{PE}_{\mathrm{sin}}$ is shown in Figure \ref{fig:sin-peak-embd}. 

\subsubsection{Transformer}

The set of fragment peaks and the precursor are passed through either of the two peak embeddings specified above. The result is a sequence of embedding vectors which is then passed through a transformer encoder: 
\begin{multline}
\mathrm{SpectrumEncoder}\left(S\right)=\\
\mathrm{TransformerEncoder}\left(\mathrm{PE}\left(m/z,I\right)_{\mathrm{precursor}},\ldots,\mathrm{PE}\left(m/z,I\right)_{\mathrm{fragment}_{N}}\right).\label{eq:SpectrumEncoder-def}
\end{multline}
Here $\mathrm{PE}$ stands in for either $\mathrm{PE}_{\mathrm{token}}$ or $\mathrm{PE}_{\mathrm{sin}}$. $\mathrm{TransformerEncoder}$ has embedding dimension $d=512$ and six layers, each with 32 attention heads and an inner hidden dimension of $d$. Note that all hidden and embedding layers in our peak embeddings above have dimension $d$ as well. As mass spectra have no intrinsic ordering, we opt to not include a positional encoding.

\begin{figure}
\begin{centering}
\includegraphics[scale=0.4]{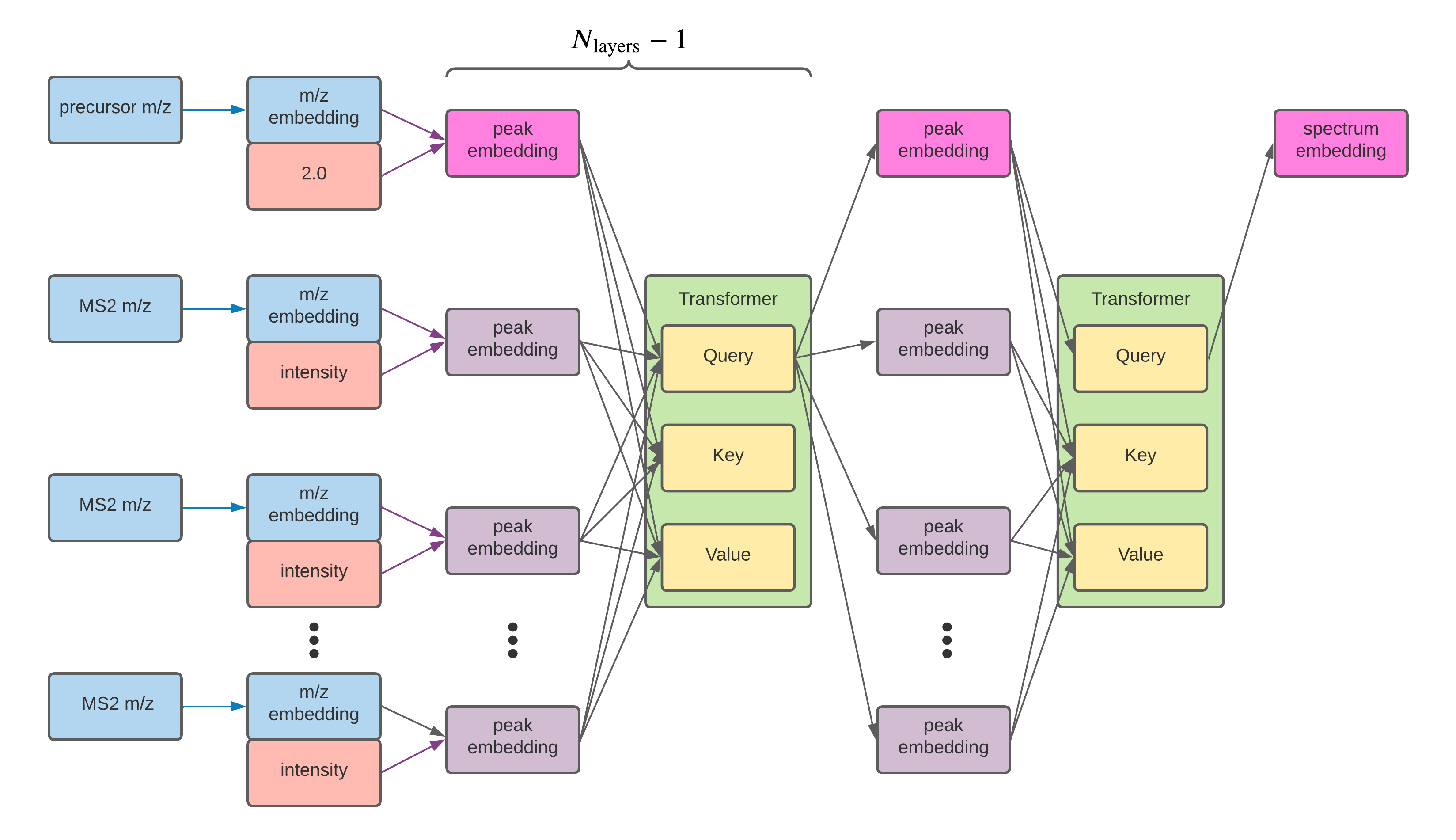} 
\par\end{centering}
\caption{\label{fig:transformer} Full model architecture that highlights how we featurize and embed an MS2 spectrum in a dense vector space.}
\end{figure}

In order to get a single embedding vector as an output, the final transformer layer query only attends to the first embedding, corresponding to the position of the precursor $m/z$. A diagram of our full model architecture is shown in Figure \ref{fig:transformer}.

\subsection{Model training\label{subsec:Model-training}}

We train the base models discussed here via two independent tasks aimed at applications within metabolomics and medicinal chemistry. These tasks are described in detail below. We build these models using PyTorch \citep{falcon2019pytorch,paszke2019pytorch} and train them with the Adam \citep{kingma2014adam} optimization algorithm with parameters $\beta_{1}=0.9$, $\beta_{2}=0.999$, and learning rate $\alpha=5.0\times10^{-5}$. We also use a weight decay \citep{krogh1991simple} parameter of $0.1$, a gradient clipping a parameter of $0.5$, and a dropout \citep{srivastava2014dropout} parameter of $0.1$. All models were trained using between 25 and 50 epochs. Finally, all model weights were trained using half precision floating point values, regardless of the precision of the $m/z$ inputs.

\subsubsection{Spectral similarity with Siamese networks \label{subsec:Spectral-similarity-with}}

A common workflow for the analysis of mass spectrometry data is to compute some notion of spectral similarity for pairs of spectra that is intended to correlate with the molecular similarity of the underlying compounds \citep{yilmaz2017methods}. This can then be used for tasks such as spectral library search \citep{stein1994optimization} and molecular networking \citep{quinn2017molecular,watrous2012mass}. Historically, most spectral similarity metrics used in the field have been heuristic-based \citep{li2021spectral,yilmaz2017methods}. More recently, spectral similarity methods that make use of techniques from deep learning have been developed \citep{huber2021spec2vec,huber2021ms2deepscore}. The latter of these, MS2Deepscore, is a Siamese network that trains a simple MLP (by binning $m/z$) to predict a measure of molecular similarity from pairs of mass spectra. As is common practice in this field, we use a molecular similarity given by a Tanimoto similarity computed from RDKit \cite{landrum2013rdkit} topological fingerprints of the corresponding molecular structures \citep{bajusz2015tanimoto}.
The loss function is the mean square error ($MSE$) between the ground truth molecular similarity and the cosine similarity evaluated on the dense embedding vectors generated by the base model.

Molecular similarity scores range from $0$ to $1$. The scores of randomly sampled pairs from our labeled dataset are not uniformly distributed and strong similarity matches are rare. \citet{huber2021ms2deepscore}, describe a procedure to sample pairs of spectra such that the molecular similarities will be uniformly distributed. We use this procedure to sample pairs for training and evaluation tasks.

Following \citet{huber2021ms2deepscore} we also train a Siamese network, but we do so using the base model transformer architectures outlined in Section \ref{subsec:Base-model-architecture} instead of with the binned mass vectors used by Huber. We report performance results for models trained using $\mathrm{PE}_{\mathrm{token}}$ and $\mathrm{PE}_{\mathrm{sin}}$ below in Section \ref{subsec:Spectral-similarity}. 

\subsubsection{Property prediction\label{subsec:Property-prediction}}

We also train the base model architecture to predict 10 chemical properties relevant to medicinal chemistry directly from an MS2 spectrum. We do so by training the following network, 
\begin{equation}
\label{eq:prop_pred}
\mathrm{Properties}=\mathrm{FF}\left(\mathrm{SpectrumEncoder}\left(S\right)\right),
\end{equation}
where the final layer has dimension equal to the number of properties we wish to predict. We also scale all of our training labels to zero mean and unit variance. For inference, we run predicted properties through an inverse-scaler. As in Section \ref{subsec:Spectral-similarity-with}, we train multiple versions of this property prediction model. Due to the lack of existing methods for comparison, we also train a version of Equation \ref{eq:prop_pred} on binned spectrum representations for a simple baseline. We replace $\mathrm{SpectrumEncoder}$ with the MS2Deepscore feed forward architecture found in \citet{huber2021ms2deepscore} for this baseline.

The properties we predict are standard indicators of druglikeness and bioavailability in the field of medicinal chemistry. These include the properties that make up the Quantitative Estimate of Druglikeness (QED) along with several others \citep{bickerton2012quantifying,lipinski2012experimental,veber2002molecular}. For the complete list of properties, see Table \ref{prop-pred-table}. All properties are computed deterministically from chemical structure using RDKit \citep{landrum2013rdkit}, so there is no additional dependency on experimental data.

\section{Results\label{sec:Results}}

\subsection{Data\label{subsec:Data}}


We construct a set of labeled MS2 spectra by combining a number of standard public and commercial datasets and one small dataset ($\sim0.1\%$ of spectra) that requires permission to access because it contains psychoactive substances \citep{wang2016sharing, mehta2020massbank, sawada2012riken, horai2010massbank, mikaia2014nist, smith2005metlin, mardal2019highresnps}. See our Reproducibility Statement for additional preprocessing details. Chemical properties and fingerprints are then computed from the cleaned molecules using RDKit  \citep{landrum2013rdkit}. Our resulting dataset has $1,\!251,\!830$ spectra corresponding to $45,\!351$ distinct structures.

In metabolomics experiments, biological samples contain compounds that have previously been profiled in MS2 libraries (``known" compounds) and compounds that have not been profiled (``novel compounds"). Identifying the structure and properties of known molecules is an important task on its own (\textit{e.g.}~“dereplication”). We therefore split our data so that the test and development sets are partially disjoint from train at the level of structures, and fully disjoint from train at the level of spectra. This enables us to evaluate model performance on both ``known" and ``novel" compounds. To implement this split, we partition our set of labeled spectra into a training set, development set (\textit{i.e.}~validation set), and test set as follows. We randomly select a set of $1002$ molecules from our data and place them and all of their associated spectra into the development set. From the remaining molecules, we follow the same procedure and place $1002$ molecules into the test set. This creates a test and development set containing $1002$ molecules each that are fully disjoint from each other. The remaining molecules are assigned the training set. We then remove $998$ spectra that correspond to $941$ molecules represented in the training set to the development set, and $998$ to the test set, corresponding to 953 structures from the training set. The final training, development, and test sets have $1,\!214,\!812$, $18,\!750$, and $18,\!268$ spectra respectively. The training set contains $43,\!347$ structures. We use the development set for hyperparameter optimization and only access the test set to compute reported metrics.


\subsection{Spectral similarity\label{subsec:Spectral-similarity}}


To benchmark spectral library search, we apply two criteria for accurate molecular identification: retrieving the exact molecule and retrieving a molecule which is an approximate match, \textit{i.e.}~has molecular similarity (defined in section \ref{subsec:Spectral-similarity-with}) greater than $0.6$.
To account for widely varying numbers of spectra for each molecule, we report the accuracies as macro averages over molecular structures.


We train a number of spectral similarity models that report state of the art performance or are in standard usage as described in Section \ref{subsec:Spectral-similarity-with}. We refer to the sinusoidal Siamese transformer and tokenized Siamese transformer models as the sinusoidal and tokenized models respectively. To test performance against reported state of the art models, we train both a Spec2Vec \citep{huber2021spec2vec} and an MS2Deepscore \citep{huber2021ms2deepscore} model on our data. We also show results modified cosine spectral similarity, which is a standard, unlearned, similarity function common in applications \citep{watrous2012mass}. In Table \ref{tab:spec-sim-table} we report spectral library search accuracies for both known and novel compounds obtained from a number of spectral similarity models. We omit exact match accuracies for novel compounds since these have been sampled so that no exact matches can be found. For models where it is applicable, the same table reports the $MSE$ loss between predicted and actual molecular similarity on spectrum pairs drawn from train, known, and novel spectra sets. 

We find that learned approaches improve on modified-cosine spectral library search. Of the learned approaches, we find that Spec2Vec has strong, only surpassed by the double precision sinusoidal model, spectral library search performance on novel, but only beats MS2Deepscore on known. MS2Deepscore only outperforms the tokenized model on novel and underperforms everything on known. Moreover, MS2Deepscore is substantially less accurate at predicting the specific pair molecular similarity on known and novel spectra when compared to all transformer approaches. The double precision sinusoidal model produced the best performance on all evaluation tasks. This model produces state of the art known spectral library search accuracies of $0.937$ and $0.97$ for exact and approximate matching respectively. Numerical representation of MS2 $m/z$ via sinusoidal embeddings consistently outperforms all the discretization based approaches we benchmarked.

To assess whether the multi-scale embeddings are learning from high resolution information, we also train and evaluate versions of our sinusoidal model where $m/z$'s are cast to half precision floating point values. Half precision floating point can only resolve up to approximately parts per $\mathcal{O}(10,000)$, far below the parts per million precision of experimental mass spectrometry data. The performance of the half precision model, reported in Table \ref{tab:spec-sim-table}, drops to be on par with the tokenized model (worse on known and better on novel). We do not report results with single precision $m/z$ since these were indistinguishable from double precision $m/z$. On all evaluation metrics, using $m/z$ inputs of precision greater than 16 bits dramatically improves performance of the sinusoidal model. This indicates that sinusoidal embedding of $m/z$ values enables learning from the the high resolution portion of mass spectrometry data.

\begin{table}
\caption{\label{tab:spec-sim-table}Spectral library search performance}
\vspace{7pt}
 \centering{}%
\begin{tabular}{ccccccc}
\toprule 
 & \multicolumn{3}{c}{$MSE$} & \multicolumn{3}{c}{Spectral library search accuracy}\tabularnewline
Spectral Set  & train  & known  & novel  & \multicolumn{2}{c}{known} & \multicolumn{1}{c}{novel}\tabularnewline
Match  & \multicolumn{3}{c}{} & Exact & Approx. & Approx.\tabularnewline
\midrule 
Modified Cosine  & \multicolumn{3}{c}{} & 0.61  & 0.672  & 0.351\tabularnewline
Spec2Vec  & \multicolumn{3}{c}{} & 0.903  & 0.954  & 0.414\tabularnewline
MS2Deepscore  & 0.026  & 0.029  & 0.047  & 0.852  & 0.918  & 0.387\tabularnewline
Tokenized $m/z$  & 0.017  & 0.019  & 0.030  & 0.933  & 0.961  & 0.382\tabularnewline
Sinusoidal  & \multirow{2}{*}{0.019} & \multirow{2}{*}{0.019} & \multirow{2}{*}{0.03} & \multirow{2}{*}{0.913} & \multirow{2}{*}{0.955} & \multirow{2}{*}{0.401}\tabularnewline
$m/z$ (\texttt{float16})  &  &  &  &  &  & \tabularnewline
Sinusoidal  & \multirow{2}{*}{\textbf{0.013}} & \multirow{2}{*}{\textbf{0.015}} & \multirow{2}{*}{\textbf{0.024}} & \multirow{2}{*}{\textbf{0.937}} & \multirow{2}{*}{\textbf{0.97}} & \multirow{2}{*}{\textbf{0.435}}\tabularnewline
$m/z$ (\texttt{float64})  &  &  &  &  &  & \tabularnewline
\bottomrule
\end{tabular}
\end{table}


\subsubsection{Qualitative embedding analysis\label{subsec:umap}}

\begin{figure}
\subfloat[\textbf{\label{fig:mz_umap_a}}]{\includegraphics[width=0.99\linewidth]{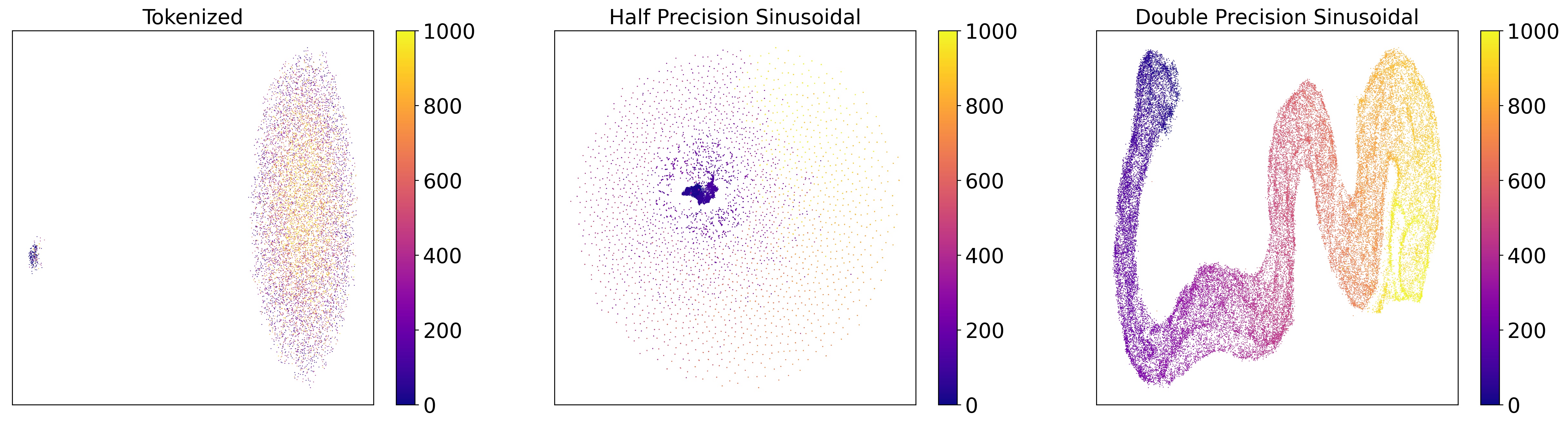} 

}

\subfloat[\label{fig:mz_umap_b}]{\includegraphics[width=0.99\linewidth]{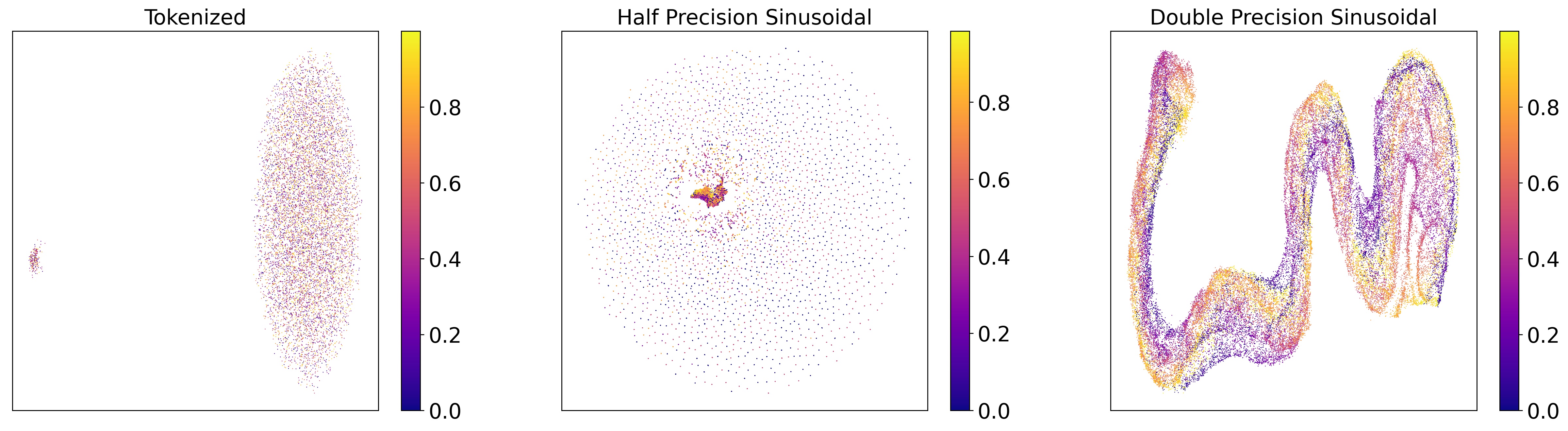} 

}

\caption{\label{fig:mz_umap} (\ref{fig:mz_umap_a}) UMAP projections of $m/z$ embeddings colored by $m/z$ value. (\ref{fig:mz_umap_b}) UMAP projections of $m/z$ embeddings colored by $\{m/z\}$. High resolution sinusoidal embeddings imbue model latent space with additional structure not found in other baseline models.}
\end{figure}

To further characterize the respective properties of token and multi-scale sinusoidal embeddings, we inspect UMAP \citep{mcinnes2018umap} projections of our $m/z$ embeddings in Figure \ref{fig:mz_umap}. For this analysis, we use siamese transformer models described in Section \ref{subsec:Spectral-similarity-with}. For our sinusoidal models, we embed $50,\!000$ $m/z$ values between $0$ and $1,\!000\,\mathrm{Daltons}$ using $\mathrm{FF}\left(\mathrm{SE}\left(m/z\right)\right)$. We do not use the full peak embedding function as we are not interested in the embedding of intensity information. Because our tokenization procedure involves rounding $m/z$ values to 1 decimal place, for the tokenized model we only embed $10,\!000$ $m/z$ values between $0$ and $1,\!000\,\mathrm{Daltons}$.

While $m/z$ measurements are continuous, the space they represent is inherently discrete. Two molecular fragments differing in atomic composition could nevertheless have a very similar $m/z$'s. However, the large separation between molecular mass scales and the mass scale at which nuclear forces manifest themselves means that the relative atomic composition may be deduced from the fractional $m/z$ \citep{jones2004strategies,pourshahian2008application}, defined by $\left\{m/z\right\}=m/z-\left\lfloor m/z\right\rfloor$.
Therefore, we expect a quality embedding of $m/z$ to embed fragments similar in both $m/z$ and $\left\{m/z\right\}$ close to one another, while paying greater attention to the latter.

As is seen in Figure \ref{fig:mz_umap}, embeddings from tokenized and low precision sinusoidal models are able to capture general trends in $m/z$. However, they fail to preserve distances in $m/z$ and show little to no structure in $\left\{m/z\right\}$. In contrast, the high precision sinusoidal embeddings allow our models to represent important information in $\left\{m/z\right\}$, while preserving distance in $m/z$. 




\subsection{Property prediction\label{subsec:Property-prediction-1}}

Many applications in metabolomics can be unblocked with knowledge of just a limited set of chemical properties, without need for identification of molecular structure. Towards this end, we investigate the parallel task of chemical property prediction from MS2 spectra. We evaluate sinusoidal \textit{vs}.~tokenized $m/z$ embedding performance on a list of properties that have compelling applications in drug discovery and are easily computable from our molecule labels. For this experiment, we train a sinusoidal model and several baseline models as described in Section \ref{subsec:Property-prediction}.

In Table \ref{prop-pred-table}, property prediction models are evaluated on known and novel structures using $R^{2}$. All transformer models outperform our feed forward baseline. Half precision sinusoidal models produced an $R^2$ of $0.746$ ($0.936$) on known (novel) molecules averaged across all 10 properties. These results are competitive with tokenization but substantially underperform when compared to higher precision sinusoidal embeddings. When $m/z$ values are specified at double precision, sinusoidal embeddings show improvements over tokenization on each property, resulting in an $11\%$ improvement over tokenization on $R^2$ for novel molecules averaged across all properties. 

This is further evidence that the multi-scale information is captured by  sinusoidal embeddings and allows improved generalization across tasks. In addition, the performance is strong enough to enable medicinal chemistry applications, such as hit prioritization based on druglikeness. This has major implications in the drug discovery context, as inspecting the properties of unknown molecules in complex mixtures was previously not possible without difficult isolation steps and individual experimentation.

\begin{table}
\caption{\label{prop-pred-table} $R^{2}$ for predicted chemical properties}
\vspace{7pt}
\centering{}%
\begin{tabular}{ccccccc}
\toprule 
\multirow{2}{*}{Property} & \multicolumn{2}{c}{Feed Forward} & \multicolumn{2}{c}{Tokenized $m/z$} & \multicolumn{2}{c}{Sinusoidal $m/z$}\tabularnewline
 & Known & Novel & Known & Novel & Known & Novel\tabularnewline
\midrule
all & 0.824 & 0.604 & 0.943 & 0.72 & \textbf{0.976} & \textbf{0.8}\tabularnewline
atomic $\log P$ & 0.768 & 0.357 & 0.919 & 0.437 & \textbf{0.968} & \textbf{0.622}\tabularnewline
number of hydrogen bond acceptors & 0.835 & 0.762 & 0.969 & 0.887 & \textbf{0.987} & \textbf{0.941}\tabularnewline
number of hydrogen bond donors & 0.834 & 0.572 & 0.924 & 0.683 & \textbf{0.968} & \textbf{0.69}\tabularnewline
polar surface area & 0.851 & 0.716 & 0.965 & 0.862 & \textbf{0.986} & \textbf{0.907}\tabularnewline
number of rotatable bonds & 0.801 & 0.669 & 0.939 & 0.792 & \textbf{0.976} & \textbf{0.838}\tabularnewline
number of aromatic rings & 0.838 & 0.372 & 0.934 & 0.474 & \textbf{0.981} & \textbf{0.655}\tabularnewline
number of aliphatic rings & 0.839 & 0.638 & 0.943 & 0.762 & \textbf{0.974} & \textbf{0.821}\tabularnewline
number of heteroatoms & 0.823 & 0.719 & 0.968 & 0.902 & \textbf{0.988} & \textbf{0.946}\tabularnewline
fraction of sp3 carbons & 0.846 & 0.618 & 0.951 & 0.723 & \textbf{0.977} & \textbf{0.821}\tabularnewline
quantitative estimate of druglikeness & 0.806 & 0.617 & 0.912 & 0.681 & \textbf{0.952} & \textbf{0.755}\tabularnewline
\bottomrule
\end{tabular}
\end{table}

\section{Conclusions\label{sec:Conclusions}}


The results presented here are the first demonstration that the sinusoidal representation of $m/z$s measured in tandem mass spectrometry experiments enables deep learning models to learn from high resolution mass data. These results include state of the art performance on spectral library search and property prediction tasks in metabolomics. The property prediction results of are sufficient quality to functionally inform and advance drug-discovery efforts. 

Moreover, we present results that compare across numerical floating point precisions for the mass inputs, and show that sinusoidal embeddings perform better when higher mass resolution data is used. Finally, we visualize high-dimensional structure of the mass embeddings that emerges only when high precision mass values are utilized. We expect that sinusoidal embeddings of $m/z$ will be a useful component of further machine learning applications with MS2 data.


\section*{Reproducibility Statement}

We train our model on a combination of free public datasets \citep{wang2016sharing, mehta2020massbank, sawada2012riken, horai2010massbank}, commercially available datasets \citep{ mikaia2014nist, smith2005metlin}, and one small dataset ($\sim0.1\%$ of all spectra)  available if a user-group of qualified researchers is joined \citep{mardal2019highresnps}. The necessary preprocessing steps are outlined in Section \ref{subsec:Data}. Additionally, we allow spectra collected across instrument types and experimental parameters including positive and negative ion mode, collision energies, etc. We constrain that spectra in our dataset have at least 5 peaks and at least 3 decimal places of $m/z$ resolution. We strip all stereochemistry from our molecular structure labels, which is a common step taken in MS2 modeling and allows for better molecule-disjoint splitting.

The model architecture is explained in detail in Section \ref{sec:Model} and all parameters used in model training are specified in Section \ref{subsec:Model-training}. We include our source code in the supplementary materials.

\bibliography{iclr2023_conference}
\bibliographystyle{iclr2023_conference}

\end{document}

%% file: math_commands.tex
\usepackage{amsmath,amsfonts,bm}

\def\eqref#1{equation~\ref{#1}}

\def\1{\bm{1}}

\DeclareMathAlphabet{\mathsfit}{\encodingdefault}{\sfdefault}{m}{sl}
\SetMathAlphabet{\mathsfit}{bold}{\encodingdefault}{\sfdefault}{bx}{n}













%% file: main.bbl
\begin{thebibliography}{42}
\providecommand{\natexlab}[1]{#1}
\providecommand{\url}[1]{\texttt{#1}}
\expandafter\ifx\csname urlstyle\endcsname\relax
  \providecommand{\doi}[1]{doi: #1}\else
  \providecommand{\doi}{doi: \begingroup \urlstyle{rm}\Url}\fi

\bibitem[Bajusz et~al.(2015)Bajusz, R{\'a}cz, and
  H{\'e}berger]{bajusz2015tanimoto}
D{\'a}vid Bajusz, Anita R{\'a}cz, and K{\'a}roly H{\'e}berger.
\newblock Why is tanimoto index an appropriate choice for fingerprint-based
  similarity calculations?
\newblock \emph{Journal of cheminformatics}, 7\penalty0 (1):\penalty0 1--13,
  2015.

\bibitem[Bickerton et~al.(2012)Bickerton, Paolini, Besnard, Muresan, and
  Hopkins]{bickerton2012quantifying}
G~Richard Bickerton, Gaia~V Paolini, J{\'e}r{\'e}my Besnard, Sorel Muresan, and
  Andrew~L Hopkins.
\newblock Quantifying the chemical beauty of drugs.
\newblock \emph{Nature chemistry}, 4\penalty0 (2):\penalty0 90--98, 2012.

\bibitem[B{\"o}cker \& Rasche(2008)B{\"o}cker and Rasche]{bocker2008towards}
Sebastian B{\"o}cker and Florian Rasche.
\newblock Towards de novo identification of metabolites by analyzing tandem
  mass spectra.
\newblock \emph{Bioinformatics}, 24\penalty0 (16):\penalty0 i49--i55, 2008.

\bibitem[Cao et~al.(2021)Cao, Guler, Tagirdzhanov, Lee, Gurevich, and
  Mohimani]{cao2021moldiscovery}
Liu Cao, Mustafa Guler, Azat Tagirdzhanov, Yi-Yuan Lee, Alexey Gurevich, and
  Hosein Mohimani.
\newblock Moldiscovery: learning mass spectrometry fragmentation of small
  molecules.
\newblock \emph{Nature Communications}, 12\penalty0 (1):\penalty0 1--13, 2021.

\bibitem[D{\"u}hrkop et~al.(2015)D{\"u}hrkop, Shen, Meusel, Rousu, and
  B{\"o}cker]{duhrkop2015searching}
Kai D{\"u}hrkop, Huibin Shen, Marvin Meusel, Juho Rousu, and Sebastian
  B{\"o}cker.
\newblock Searching molecular structure databases with tandem mass spectra
  using csi: Fingerid.
\newblock \emph{Proceedings of the National Academy of Sciences}, 112\penalty0
  (41):\penalty0 12580--12585, 2015.

\bibitem[Dunn et~al.(2013)Dunn, Erban, Weber, Creek, Brown, Breitling,
  Hankemeier, Goodacre, Neumann, Kopka, et~al.]{dunn2013mass}
Warwick~B Dunn, Alexander Erban, Ralf~JM Weber, Darren~J Creek, Marie Brown,
  Rainer Breitling, Thomas Hankemeier, Royston Goodacre, Steffen Neumann,
  Joachim Kopka, et~al.
\newblock Mass appeal: metabolite identification in mass spectrometry-focused
  untargeted metabolomics.
\newblock \emph{Metabolomics}, 9\penalty0 (1):\penalty0 44--66, 2013.

\bibitem[Falcon(2019)]{falcon2019pytorch}
William Falcon.
\newblock The pytorch lightning team.
\newblock \emph{Pytorch lightning}, 3:\penalty0 6, 2019.

\bibitem[Glish \& Vachet(2003)Glish and Vachet]{glish2003basics}
Gary~L Glish and Richard~W Vachet.
\newblock The basics of mass spectrometry in the twenty-first century.
\newblock \emph{Nature reviews drug discovery}, 2\penalty0 (2):\penalty0
  140--150, 2003.

\bibitem[Horai et~al.(2010)Horai, Arita, Kanaya, Nihei, Ikeda, Suwa, Ojima,
  Tanaka, Tanaka, Aoshima, et~al.]{horai2010massbank}
Hisayuki Horai, Masanori Arita, Shigehiko Kanaya, Yoshito Nihei, Tasuku Ikeda,
  Kazuhiro Suwa, Yuya Ojima, Kenichi Tanaka, Satoshi Tanaka, Ken Aoshima,
  et~al.
\newblock Massbank: a public repository for sharing mass spectral data for life
  sciences.
\newblock \emph{Journal of mass spectrometry}, 45\penalty0 (7):\penalty0
  703--714, 2010.

\bibitem[Huber et~al.(2021{\natexlab{a}})Huber, Ridder, Verhoeven, Spaaks,
  Diblen, Rogers, and Van Der~Hooft]{huber2021spec2vec}
Florian Huber, Lars Ridder, Stefan Verhoeven, Jurriaan~H Spaaks, Faruk Diblen,
  Simon Rogers, and Justin~JJ Van Der~Hooft.
\newblock Spec2vec: Improved mass spectral similarity scoring through learning
  of structural relationships.
\newblock \emph{PLoS computational biology}, 17\penalty0 (2):\penalty0
  e1008724, 2021{\natexlab{a}}.

\bibitem[Huber et~al.(2021{\natexlab{b}})Huber, van~der Burg, van~der Hooft,
  and Ridder]{huber2021ms2deepscore}
Florian Huber, Sven van~der Burg, Justin~JJ van~der Hooft, and Lars Ridder.
\newblock Ms2deepscore: a novel deep learning similarity measure to compare
  tandem mass spectra.
\newblock \emph{Journal of cheminformatics}, 13\penalty0 (1):\penalty0 1--14,
  2021{\natexlab{b}}.

\bibitem[Jones et~al.(2004)Jones, Stump, Fleming, Lay, and
  Wilkins]{jones2004strategies}
Jeffrey~J Jones, Michael~J Stump, Richard~C Fleming, Jackson~O Lay, and
  Charles~L Wilkins.
\newblock Strategies and data analysis techniques for lipid and phospholipid
  chemistry elucidation by intact cell maldi-ftms.
\newblock \emph{Journal of the American Society for Mass Spectrometry},
  15\penalty0 (11):\penalty0 1665--1674, 2004.

\bibitem[Kingma \& Ba(2014)Kingma and Ba]{kingma2014adam}
Diederik~P Kingma and Jimmy Ba.
\newblock Adam: A method for stochastic optimization.
\newblock \emph{arXiv preprint arXiv:1412.6980}, 2014.

\bibitem[Krogh \& Hertz(1991)Krogh and Hertz]{krogh1991simple}
Anders Krogh and John Hertz.
\newblock A simple weight decay can improve generalization.
\newblock \emph{Advances in neural information processing systems}, 4, 1991.

\bibitem[Kutuzova et~al.(2021)Kutuzova, Igel, Nielsen, and
  McCloskey]{kutuzova2021bi}
Svetlana Kutuzova, Christian Igel, Mads Nielsen, and Douglas McCloskey.
\newblock Bi-modal variational autoencoders for metabolite identification using
  tandem mass spectrometry.
\newblock \emph{bioRxiv}, 2021.

\bibitem[Landrum et~al.(2013)]{landrum2013rdkit}
Greg Landrum et~al.
\newblock Rdkit: A software suite for cheminformatics, computational chemistry,
  and predictive modeling.
\newblock \emph{Greg Landrum}, 2013.

\bibitem[Li et~al.(2021{\natexlab{a}})Li, Si, Li, Hsieh, and
  Bengio]{li2021learnable}
Yang Li, Si~Si, Gang Li, Cho-Jui Hsieh, and Samy Bengio.
\newblock Learnable fourier features for multi-dimensional spatial positional
  encoding.
\newblock \emph{Advances in Neural Information Processing Systems},
  34:\penalty0 15816--15829, 2021{\natexlab{a}}.

\bibitem[Li et~al.(2021{\natexlab{b}})Li, Kind, Folz, Vaniya, Mehta, and
  Fiehn]{li2021spectral}
Yuanyue Li, Tobias Kind, Jacob Folz, Arpana Vaniya, Sajjan~Singh Mehta, and
  Oliver Fiehn.
\newblock Spectral entropy outperforms ms/ms dot product similarity for
  small-molecule compound identification.
\newblock \emph{Nature Methods}, 18\penalty0 (12):\penalty0 1524--1531,
  2021{\natexlab{b}}.

\bibitem[Lipinski et~al.(2012)Lipinski, Lombardo, Dominy, and
  Feeney]{lipinski2012experimental}
Christopher~A Lipinski, Franco Lombardo, Beryl~W Dominy, and Paul~J Feeney.
\newblock Experimental and computational approaches to estimate solubility and
  permeability in drug discovery and development settings.
\newblock \emph{Advanced drug delivery reviews}, 64:\penalty0 4--17, 2012.

\bibitem[Litsa et~al.(2021)Litsa, Chenthamarakshan, Das, and
  Kavraki]{litsa2021spec2mol}
Eleni Litsa, Vijil Chenthamarakshan, Payel Das, and Lydia Kavraki.
\newblock Spec2mol: An end-to-end deep learning framework for translating ms/ms
  spectra to de-novo molecules.
\newblock 2021.

\bibitem[Mardal et~al.(2019)Mardal, Andreasen, Mollerup, Stockham, Telving,
  Thomaidis, Diamanti, Linnet, and Dalsgaard]{mardal2019highresnps}
Marie Mardal, Mette~Findal Andreasen, Christian~Brinch Mollerup, Peter
  Stockham, Rasmus Telving, Nikolaos~S Thomaidis, Konstantina~S Diamanti,
  Kristian Linnet, and Petur~Weihe Dalsgaard.
\newblock Highresnps. com: an online crowd-sourced hr-ms database for suspect
  and non-targeted screening of new psychoactive substances.
\newblock \emph{Journal of Analytical Toxicology}, 2019.

\bibitem[McInnes et~al.(2018)McInnes, Healy, and Melville]{mcinnes2018umap}
Leland McInnes, John Healy, and James Melville.
\newblock Umap: Uniform manifold approximation and projection for dimension
  reduction.
\newblock \emph{arXiv preprint arXiv:1802.03426}, 2018.

\bibitem[Mehta(2020)]{mehta2020massbank}
SS~Mehta.
\newblock Massbank of north america (mona): An open-access, auto-curating mass
  spectral database for compound identification in metabolomics presentation,
  2020.

\bibitem[Mikaia et~al.(2014)Mikaia, EI, EI, EI, EI, Neta, RI, Linstrom,
  Mirokhin, Tchekhovskoi, et~al.]{mikaia2014nist}
Anzor Mikaia, Principal Edward White~V EI, Vladimir~Zaikin EI, Damo~Zhu EI,
  O~David~Sparkman EI, Pedatsur Neta, Igor~Zenkevich RI, Peter Linstrom, Yuri
  Mirokhin, Dmitrii Tchekhovskoi, et~al.
\newblock Nist standard reference database 1a.
\newblock \emph{Standard Reference Data, NIST, Gaithersburg, MD, USA
  https://www. nist. gov/srd/nist-standard-reference-database-1a}, 2014.

\bibitem[Mikolov et~al.(2013)Mikolov, Chen, Corrado, and
  Dean]{mikolov2013efficient}
Tomas Mikolov, Kai Chen, Greg Corrado, and Jeffrey Dean.
\newblock Efficient estimation of word representations in vector space.
\newblock \emph{arXiv preprint arXiv:1301.3781}, 2013.

\bibitem[Paszke et~al.(2019)Paszke, Gross, Massa, Lerer, Bradbury, Chanan,
  Killeen, Lin, Gimelshein, Antiga, et~al.]{paszke2019pytorch}
Adam Paszke, Sam Gross, Francisco Massa, Adam Lerer, James Bradbury, Gregory
  Chanan, Trevor Killeen, Zeming Lin, Natalia Gimelshein, Luca Antiga, et~al.
\newblock Pytorch: An imperative style, high-performance deep learning library.
\newblock \emph{Advances in neural information processing systems}, 32, 2019.

\bibitem[Pourshahian \& Limbach(2008)Pourshahian and
  Limbach]{pourshahian2008application}
Soheil Pourshahian and Patrick~A Limbach.
\newblock Application of fractional mass for the identification of
  peptide--oligonucleotide cross-links by mass spectrometry.
\newblock \emph{Journal of mass spectrometry}, 43\penalty0 (8):\penalty0
  1081--1088, 2008.

\bibitem[Qiao et~al.(2019)Qiao, Tran, Xin, Shan, Li, and
  Ghodsi]{qiao2019deepnovov2}
Rui Qiao, Ngoc~Hieu Tran, Lei Xin, Baozhen Shan, Ming Li, and Ali Ghodsi.
\newblock Deepnovov2: Better de novo peptide sequencing with deep learning.
\newblock \emph{arXiv preprint arXiv:1904.08514}, 2019.

\bibitem[Quinn et~al.(2017)Quinn, Nothias, Vining, Meehan, Esquenazi, and
  Dorrestein]{quinn2017molecular}
Robert~A Quinn, Louis-Felix Nothias, Oliver Vining, Michael Meehan, Eduardo
  Esquenazi, and Pieter~C Dorrestein.
\newblock Molecular networking as a drug discovery, drug metabolism, and
  precision medicine strategy.
\newblock \emph{Trends in pharmacological sciences}, 38\penalty0 (2):\penalty0
  143--154, 2017.

\bibitem[Sawada et~al.(2012)Sawada, Nakabayashi, Yamada, Suzuki, Sato, Sakata,
  Akiyama, Sakurai, Matsuda, Aoki, et~al.]{sawada2012riken}
Yuji Sawada, Ryo Nakabayashi, Yutaka Yamada, Makoto Suzuki, Muneo Sato, Akane
  Sakata, Kenji Akiyama, Tetsuya Sakurai, Fumio Matsuda, Toshio Aoki, et~al.
\newblock Riken tandem mass spectral database (respect) for phytochemicals: a
  plant-specific ms/ms-based data resource and database.
\newblock \emph{Phytochemistry}, 82:\penalty0 38--45, 2012.

\bibitem[Shrivastava et~al.(2021)Shrivastava, Swainston, Samanta, Roberts,
  Wright~Muelas, and Kell]{shrivastava2021massgenie}
Aditya~Divyakant Shrivastava, Neil Swainston, Soumitra Samanta, Ivayla Roberts,
  Marina Wright~Muelas, and Douglas~B Kell.
\newblock Massgenie: A transformer-based deep learning method for identifying
  small molecules from their mass spectra.
\newblock \emph{Biomolecules}, 11\penalty0 (12):\penalty0 1793, 2021.

\bibitem[Smith et~al.(2005)Smith, O'Maille, Want, Qin, Trauger, Brandon,
  Custodio, Abagyan, and Siuzdak]{smith2005metlin}
Colin~A Smith, Grace O'Maille, Elizabeth~J Want, Chuan Qin, Sunia~A Trauger,
  Theodore~R Brandon, Darlene~E Custodio, Ruben Abagyan, and Gary Siuzdak.
\newblock Metlin: a metabolite mass spectral database.
\newblock \emph{Therapeutic drug monitoring}, 27\penalty0 (6):\penalty0
  747--751, 2005.

\bibitem[Srivastava et~al.(2014)Srivastava, Hinton, Krizhevsky, Sutskever, and
  Salakhutdinov]{srivastava2014dropout}
Nitish Srivastava, Geoffrey Hinton, Alex Krizhevsky, Ilya Sutskever, and Ruslan
  Salakhutdinov.
\newblock Dropout: a simple way to prevent neural networks from overfitting.
\newblock \emph{The journal of machine learning research}, 15\penalty0
  (1):\penalty0 1929--1958, 2014.

\bibitem[Stein \& Scott(1994)Stein and Scott]{stein1994optimization}
Stephen~E Stein and Donald~R Scott.
\newblock Optimization and testing of mass spectral library search algorithms
  for compound identification.
\newblock \emph{Journal of the American Society for Mass Spectrometry},
  5\penalty0 (9):\penalty0 859--866, 1994.

\bibitem[van Der~Hooft et~al.(2016)van Der~Hooft, Wandy, Barrett, Burgess, and
  Rogers]{van2016topic}
Justin Johan~Jozias van Der~Hooft, Joe Wandy, Michael~P Barrett, Karl~EV
  Burgess, and Simon Rogers.
\newblock Topic modeling for untargeted substructure exploration in
  metabolomics.
\newblock \emph{Proceedings of the National Academy of Sciences}, 113\penalty0
  (48):\penalty0 13738--13743, 2016.

\bibitem[Vaswani et~al.(2017)Vaswani, Shazeer, Parmar, Uszkoreit, Jones, Gomez,
  Kaiser, and Polosukhin]{vaswani2017attention}
Ashish Vaswani, Noam Shazeer, Niki Parmar, Jakob Uszkoreit, Llion Jones,
  Aidan~N Gomez, {\L}ukasz Kaiser, and Illia Polosukhin.
\newblock Attention is all you need.
\newblock \emph{Advances in neural information processing systems}, 30, 2017.

\bibitem[Veber et~al.(2002)Veber, Johnson, Cheng, Smith, Ward, and
  Kopple]{veber2002molecular}
Daniel~F Veber, Stephen~R Johnson, Hung-Yuan Cheng, Brian~R Smith, Keith~W
  Ward, and Kenneth~D Kopple.
\newblock Molecular properties that influence the oral bioavailability of drug
  candidates.
\newblock \emph{Journal of medicinal chemistry}, 45\penalty0 (12):\penalty0
  2615--2623, 2002.

\bibitem[Wang et~al.(2016)Wang, Carver, Phelan, Sanchez, Garg, Peng, Nguyen,
  Watrous, Kapono, Luzzatto-Knaan, et~al.]{wang2016sharing}
Mingxun Wang, Jeremy~J Carver, Vanessa~V Phelan, Laura~M Sanchez, Neha Garg,
  Yao Peng, Don~Duy Nguyen, Jeramie Watrous, Clifford~A Kapono, Tal
  Luzzatto-Knaan, et~al.
\newblock Sharing and community curation of mass spectrometry data with global
  natural products social molecular networking.
\newblock \emph{Nature biotechnology}, 34\penalty0 (8):\penalty0 828--837,
  2016.

\bibitem[Watrous et~al.(2012)Watrous, Roach, Alexandrov, Heath, Yang, Kersten,
  van~der Voort, Pogliano, Gross, Raaijmakers, et~al.]{watrous2012mass}
Jeramie Watrous, Patrick Roach, Theodore Alexandrov, Brandi~S Heath, Jane~Y
  Yang, Roland~D Kersten, Menno van~der Voort, Kit Pogliano, Harald Gross,
  Jos~M Raaijmakers, et~al.
\newblock Mass spectral molecular networking of living microbial colonies.
\newblock \emph{Proceedings of the National Academy of Sciences}, 109\penalty0
  (26):\penalty0 E1743--E1752, 2012.

\bibitem[Yilmaz et~al.(2022)Yilmaz, Fondrie, Bittremieux, Oh, and
  Noble]{yilmaz2022novo}
Melih Yilmaz, William Fondrie, Wout Bittremieux, Sewoong Oh, and William~S
  Noble.
\newblock De novo mass spectrometry peptide sequencing with a transformer
  model.
\newblock In \emph{International Conference on Machine Learning}, pp.\
  25514--25522. PMLR, 2022.

\bibitem[Yilmaz et~al.(2017)Yilmaz, Vandermarliere, and
  Martens]{yilmaz2017methods}
{\c{S}}ule Yilmaz, Elien Vandermarliere, and Lennart Martens.
\newblock Methods to calculate spectrum similarity.
\newblock In \emph{Proteome bioinformatics}, pp.\  75--100. Springer, 2017.

\bibitem[Zhang et~al.(2020)Zhang, Li, Dou, et~al.]{zhang2020mass}
Xi-wu Zhang, Qiu-han Li, Jin-jin Dou, et~al.
\newblock Mass spectrometry-based metabolomics in health and medical science: A
  systematic review.
\newblock \emph{RSC advances}, 10\penalty0 (6):\penalty0 3092--3104, 2020.

\end{thebibliography}
